\documentclass{article}

\usepackage[english]{babel}

\usepackage[letterpaper,top=2cm,bottom=2cm,left=3cm,right=3cm,marginparwidth=1.75cm]{geometry}
\usepackage{tabularx}
\usepackage{amsmath}
\usepackage{graphicx}
\usepackage[colorlinks=true, allcolors=blue]{hyperref}
\usepackage{longtable}
\usepackage{array}
\usepackage{pdflscape}
\usepackage{caption}
\usepackage{authblk}
\usepackage{float}
\usepackage[utf8]{inputenc}
\usepackage{array}
\usepackage{float} 
\usepackage{caption}
\usepackage{subcaption}
\title{Evaluating the Performance of Large Language Models for SDG Mapping (Technical Report)}

\author[1]{Hui Yin\thanks{* huiyin@swin.edu.au}}
\author[1]{Amir Aryani}
\author[1]{Nakul Nambiar}

\affil[1]{Swinburne University of Technology, Australia}

\begin{document}
\maketitle

\begin{abstract}
The use of large language models (LLMs) is expanding rapidly, and open-source versions are becoming available, offering users safer and more adaptable options. These models enable users to protect data privacy by eliminating the need to provide data to third parties and can be customized for specific tasks.
In this study, we compare the performance of various language models on the Sustainable Development Goal (SDG) mapping task, using the output of GPT-4o as the baseline. The selected open-source models for comparison include Mixtral, LLaMA 2, LLaMA 3, Gemma, and Qwen2. Additionally, GPT-4o-mini, a more specialized version of GPT-4o, was included to extend the comparison. Given the multi-label nature of the SDG mapping task, we employed metrics such as F1 score, precision, and recall with micro-averaging to evaluate different aspects of the models' performance. These metrics are derived from the confusion matrix to ensure a comprehensive evaluation. We provide a clear observation and analysis of each model's performance by plotting curves based on F1 score, precision, and recall at different thresholds. According to the results of this experiment, LLaMA 2 and Gemma still have significant room for improvement. The other four models do not exhibit particularly large differences in performance.
The outputs from all seven models are available on Zenodo: \url{https://doi.org/10.5281/zenodo.12789375}.


\end{abstract}

\section{Introduction}
LLMs are AI models that aim to generate human-like text by learning from vast amounts of existing data, such as Wikipedia, books, articles, and web pages. These models are based on deep neural network architectures, particularly the transformer structure, which enables them to handle large datasets and complex language tasks effectively. LLMs have been known as game-changers in Natural Language Processing (NLP) because they allow machines to perform tasks that require the comprehension and generation of highly fluent and accurate human language.
Some well-known companies' large language models are not publicly available or require subscription fees. For example, OpenAI's ChatGPT (Chat Generative Pre-Trained Transformer)\footnote{\url{https://openai.com/chatgpt/}}, Google's LaMDA\footnote{\url{https://blog.google/technology/ai/lamda/}} (Language Model for Dialogue Applications), and Microsoft's Turing-NLG (Turing Natural Language Generation)\footnote{\url{https://www.microsoft.com/en-us/research/blog/turing-nlg-a-17-billion-parameter-language-model-by-microsoft/}}.
OpenAI's ChatGPT (GPT-3.5) offers limited functionality for ordinary users, while those who pay a subscription fee can access more advanced models with richer features. For example, ChatGPT (GPT-4.0) can directly upload pictures and documents for enhanced outputs. 
Google's LaMDA is primarily used in Google's internal projects and products, and  Microsoft's Turing-NLG is used within Microsoft's suite of products and services; it is not available as an open-source model.

Uploading data to external companies presents significant security risks for confidential materials related to scientific research institutions, businesses, the military, and government entities. Therefore, open-source large language models that can run locally without exposing data to third parties have become urgent. This approach ensures data security and allows users to tailor language models for specific tasks. The excellent performance of these locally run models underscores their potential for a wide range of applications while maintaining user control and privacy. 
How do free, open-source large language models (LLMs) compare to paid models like GPT-4o? In this study, we conducted a case study using 1,000 publications from the Swinburne University of Technology research bank. Each model was given the same prompt to perform the Sustainable Development Goals (SDG) mapping task, which involved assigning each publication to different goals and providing a confidence level and an explanation for each assignment. The open-source LLMs evaluated in this experiment were Mixtral, LLaMA 2, LLaMA 3, Gemma, and Qwen2. The paid models were GPT-4o and GPT-4o-mini. Given the lack of ground truth labels in the experimental data, the output of GPT-4o was used as the baseline.
We hope the results of this experiment will help other researchers select suitable open-source LLMs for their projects.

\section{Experiment}
The experimental prototype environment is configured as follows: all open-source LLM models were executed on GPU compute nodes within the Swinburne High-Performance Computing (HPC) facility. Each node is equipped with an NVIDIA A100 GPU with 80 GB of memory, 15 GB of RAM, and a single CPU core.

To compare the performance of various language models, we used the output of GPT-4o as the baseline. We selected the following open-source models for comparison: Mixtral, LLaMA 2, LLaMA 3, Gemma 2, and Qwen2. Additionally, we included GPT-4o-mini, a smaller version of GPT-4o, to extend the comparison.
Each model is briefly described as follows:
\begin{itemize}
    \item GPT-4o\footnote{\url{https://openai.com/index/hello-gpt-4o/}}: The new flagship model that can reason across audio, vision, and text in real-time. It matches GPT-4 Turbo performance on text in English and code, with significant improvement on text in non-English languages, while also being much faster and 50\% cheaper in the API.
    \item GPT-4o-mini\footnote{\url{https://openai.com/index/gpt-4o-mini-advancing-cost-efficient-intelligence/}}: A smaller variant of GPT-4o designed for efficient performance with fewer computational resources. Details on parameters and architecture are proprietary.
    \item Mixtral\footnote{\url{https://huggingface.co/docs/transformers/en/model_doc/mixtral}}: An open-source language model with various versions, designed for versatile applications in natural language processing. Exact parameter counts and training data specifics vary across versions.
    \item LLaMA 2\footnote{\url{https://llama.meta.com/llama2/}}: A collection of pre-trained and fine-tuned text models ranging in scale from 7 billion to 70 billion parameters, designed for a variety of natural language understanding and generation tasks.
    \item LLaMA 3\footnote{\url{https://llama.meta.com/llama3/}}: A collection of pre-trained and fine-tuned text models with two sizes: 8 billion and 70 billion parameters, pre-trained on 15 trillion tokens. LLaMA 3 is an advancement over LLaMA 2, providing improved performance and efficiency.
    \item Gemma 2\footnote{\url{https://storage.googleapis.com/deepmind-media/gemma/gemma-2-report.pdf}}: A new addition to the Gemma family of lightweight, ranging in scale from 9 billion to 27 billion parameters are available today. These models are optimized for both performance and efficiency in various natural language processing tasks.
    \item Qwen2\footnote{\url{https://huggingface.co/Qwen}}: An open-source language model designed for general-purpose applications in natural language processing. The specific parameter counts and architectural details are tailored to balance performance with resource efficiency.
\end{itemize}

In this study, we randomly selected 1,000 publications from the Swinburne University of Technology research bank, using their titles and abstracts as input. 
We employed the seven models to perform the SDG mapping task, which involves 17 closely interconnected Sustainable Development Goals (SDGs)\footnote{\url{https://sdgs.un.org/goals}}, resulting in a single publication potentially being relevant to multiple goals.
The designed prompt for analysis was: PROMPT = ```Analyze the publication and determine the SDGs it aligns with. Evaluate against all the 17 SDGs and provide the reason for alignment. In the end summarise the confidence levels(\%) for each assigned SDG in JSON format. For example: {example}. Title: {title} Description: {description}''', example = ```\{`Goal 6': 0.67, `Goal 11': 0.50, `Goal 3': 0.25\}'''.
To ensure a consistent evaluation, we used the same prompt for all models, thus standardizing the difficulty level across them. 
The designed prompts ensure that all publications are assessed against all 17 SDGs and provide explanations for each alignment. There is no limit on the number of tokens, resulting in each output containing a large amount of text. 
Table~\ref{tab_GPT-4o_Output} presents the output from GPT-4o for the publication ``IRS-aided Received Signal Strength Localization Using a Wireless Sensor Network\footnote{\url{https://www.researchgate.net/publication/378758478_IRS-aided_Received_Signal_Strength_Localization_Using_a_Wireless_Sensor_Network}}” as a demonstration. 
\begin{table}[]
\caption{Sample output from GPT-4o for the SDG mapping task on a selected publication.}
\label{tab_GPT-4o_Output}
\begin{tabularx}{\textwidth}{|p{0.1\textwidth}|X|}
\hline
\textbf{Category} & \textbf{Content} \\                                                      \hline
Title & IRS-aided Received Signal Strength Localization Using a Wireless Sensor Network  \\ \hline
Abstract & In this letter, we initially present a formulation of Received Signal Strengths (RSS) in a scenario involving one source (or emitter), one Intelligent Reflecting Surface (IRS), and a Wireless Sensor Network (WSN). To determine the source location based on RSS measurements, we carefully select IRS phases to simplify the relationship between them. Subsequently, we solve the resulting rather simple nonlinear equations using a Least Squares (LS) approach. The solution involves an initial estimation through a course search, followed by a Steepest-Descent (SD) recursion. Simulation results highlight the superior performance of SD compared to course search and some state-of-the-art RSS localization techniques in the literature. \\ \hline

GPT-4o Output & objectives of the SDGs. Here's an evaluation against all 17 SDGs:

1. **No Poverty**: This publication does not directly address poverty alleviation.
2. **Zero Hunger**: The research does not directly relate to improving food security or agricultural productivity.
3. **Good Health and Well-being**: There is no explicit mention of direct implications for health and well-being.
4. **Quality Education**: The content does not directly relate to improving the quality or access to education.
5. **Gender Equality**: It does not address issues related to gender equality.
6. **Clean Water and Sanitation**: The research does not have a direct impact on water or sanitation issues.
7. **Affordable and Clean Energy**: The publication does not discuss energy issues.
8. **Decent Work and Economic Growth**: While the research may have indirect implications for technological innovation, it does not explicitly focus on employment or economic growth.
9. **Industry, Innovation, and Infrastructure**: This publication fits well here as it discusses innovative techniques in wireless sensor networks. There is a strong alignment due to the focus on improving technology infrastructure.
10. **Reduced Inequalities**: The scope does not address social or economic inequalities directly.
11. **Sustainable Cities and Communities**: Enhanced localization techniques could indirectly support smart city applications, making cities more efficient.
12. **Responsible Consumption and Production**: There is no clear linkage to responsible consumption or production.
13. **Climate Action**: No direct mention or implication for climate action is present.
14. **Life Below Water**: The publication does not address issues related to marine life.
15. **Life on Land**: It does not focus on terrestrial ecosystems.
16. **Peace and Justice Strong Institutions**: It does not address issues related to peace, justice, or institutions.
17. **Partnerships for the Goals**: The publication might indirectly contribute to partnerships through technological advancements but this association is tenuous.

**Summary of Confidence Levels:**
- **Goal 9 (Industry, Innovation, and Infrastructure)**: There is a high confidence level in aligning with this goal due to the focus on enhancing wireless sensor networks and localization techniques.
- **Goal 11 (Sustainable Cities and Communities)**: There is moderate confidence as improved RSS localization can contribute to smart city technologies and urban planning.
```json
{
    "Goal 9": 0.8,
    "Goal 11": 0.5
}
```
In summary, the highest confidence is in Goal 9 due to the publication's direct focus on technological innovations that can bolster industry and infrastructure. Goal 11 has a moderate level of alignment through the potential application of these technologies in creating more sustainable urban environments.
\\ \hline
                                         
\end{tabularx}
\end{table}

From table~\ref{tab_GPT-4o_Output}, we can see that the output is generated according to the prompts. It first compares all 17 goals, then provides a conclusion with an explanation and confidence level, and finally presents the answer in JSON format based on the provided example. 
The outputs from other models vary significantly in their conclusions and text formats. For complete results, please refer to the dataset uploaded to Zenodo.
Table~\ref{tab_model_parameter_com} shows some data related to the model operation.
Only GPT-4o and GPT-4o-mini require payment to OpenAI, while the other five models are open-source and free. For processing the prompt in the SDG mapping task across 1,000 publications, the cost of GPT-4o is 32 times higher than that of GPT-4o-mini. Additionally, GPT-4o-mini significantly reduces processing time compared to GPT-4o, which is the fastest among the 7 models; Qwen2 took the longest, 7 hours and 41 minutes.
The practical implementation varied depending on the physical size of each model. Larger models were challenging to load into memory and required significantly more time, while smaller models loaded quickly and were easier to manage.  

\begin{table}[H]
\centering
\begin{tabular}{|m{1.4 cm}|m{1.7cm}|m{1.5 cm}|m{1.5cm}|m{1.5cm}|m{1.5cm}|m{1.5cm}|m{1.5cm}|}
\hline
 \centering Model        & GPT-4o &  \centering GPT-4o-mini &  \centering Mixtral 8x22B &  \centering LLaMA 2 70B &  \centering LLaMA 3 70B &  \centering Gemma 2 27B & Qwen2  72B \\ \hline
 \centering Running Time & 4h 33m & 2h 49m & 5h 18m & 4h 58m & 6h 1m & 5h 50m & 7h 41m \\ \hline
 \centering Cost         & \centering14.58 USD  &   0.46 USD  & - & - & - & - & - \\ \hline
\end{tabular}
\caption{Model comparison based on running time and cost. The configurations (e.g., 8x22B, 70B, 27B) indicate the parameter settings used for each model in the experiment.}
\label{tab_model_parameter_com}
\end{table}

\section{Result Comparison}
Table~\ref {tab_GPT-4o_Output} shows that the output of the large language model for SDG mapping is presented as a whole paragraph of text. While some models also provide results in the JSON format specified by the prompt, this format is not conducive to our performance comparison between models. Therefore, we processed the results to extract only the goals and confidence scores for comparison. 
The processed output, such as ``Goal 9: 0.85, Goal 11: 0.8, Goal 12: 0.6, Goal 3: 0.3, Goal 7: 0.2", allows us to compare model outputs, each goal corresponds to a confidence level. 
To evaluate the models, we set thresholds at intervals of 0.1 (e.g., 0.1, 0.2, 0.3, etc.). 
We use the output of GPT-4o (with a threshold of 0.5) as the baseline to assess the performance of the other models.
This method enables us to evaluate and compare the models' performance based on different confidence levels, helping to identify which models provide more consistent and reliable mappings to the SDGs.
We employed the F1 score, precision, and recall from the confusion matrix to measure the models' performance because SDG mapping is a multi-label classification task where a publication can be assigned to multiple goals. 
Micro-averaged metrics were chosen to provide an overall performance metric that reflects the effectiveness of the models across all instances. This approach helps understand the general robustness of the models. 
To compare each model's performance against GPT-4o, we plotted curves based on the F1 score, precision, and recall values obtained at different thresholds. These visualizations allow for more precise observation and analysis of how each model performs across various confidence levels, providing insights into their effectiveness and reliability in the SDG mapping task.

\begin{figure}
    \centering
    \includegraphics[width=0.7\linewidth]{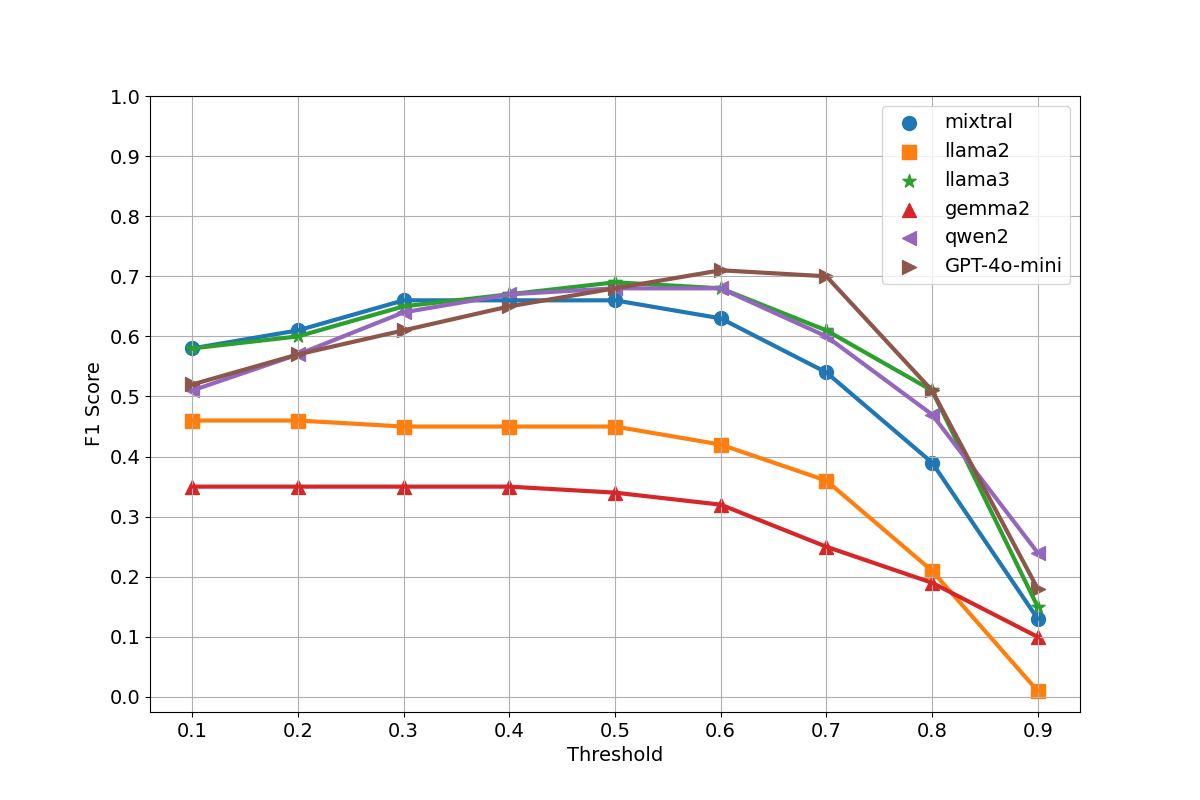}
    \caption{Comparison of F1 scores across different model thresholds.}
    \label{fig:enter-label}
\end{figure}

F1 Score is generally the most informative single metric in multi-label classification, balancing precision and recall.
The top-performing models across most thresholds are GPT-4o-mini, LLaMA 3, and Qwen2, maintaining higher F1 scores consistently. 
Gemma 2 and LLaMA 2 consistently show the lowest F1 scores among the models, suggesting they are ineffective at any threshold value tested; LLaMA 2 is slightly better than Gemma 2.
The optimal threshold range for most models (Mixtral, LLaMA3, Qwen2, and GPT-4o-mini) appears to be between 0.5 and 0.6, where they achieve their highest F1 scores.

\begin{figure}
     \centering
     \begin{subfigure}[b]{0.48\textwidth}
         \centering
         \includegraphics[width=\textwidth]{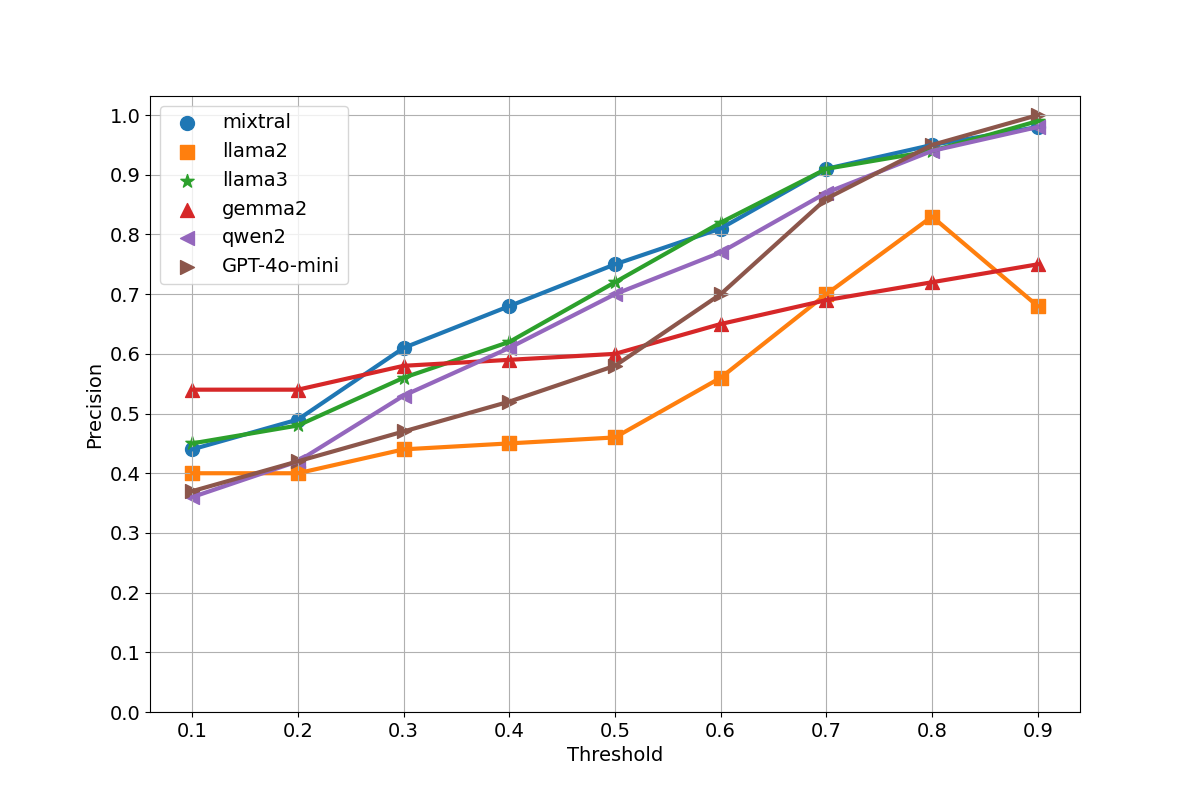}
         \caption{Precision}
         \label{fig:precision}
     \end{subfigure}
     \hfill
     \begin{subfigure}[b]{0.48\textwidth}
         \centering
         \includegraphics[width=\textwidth]{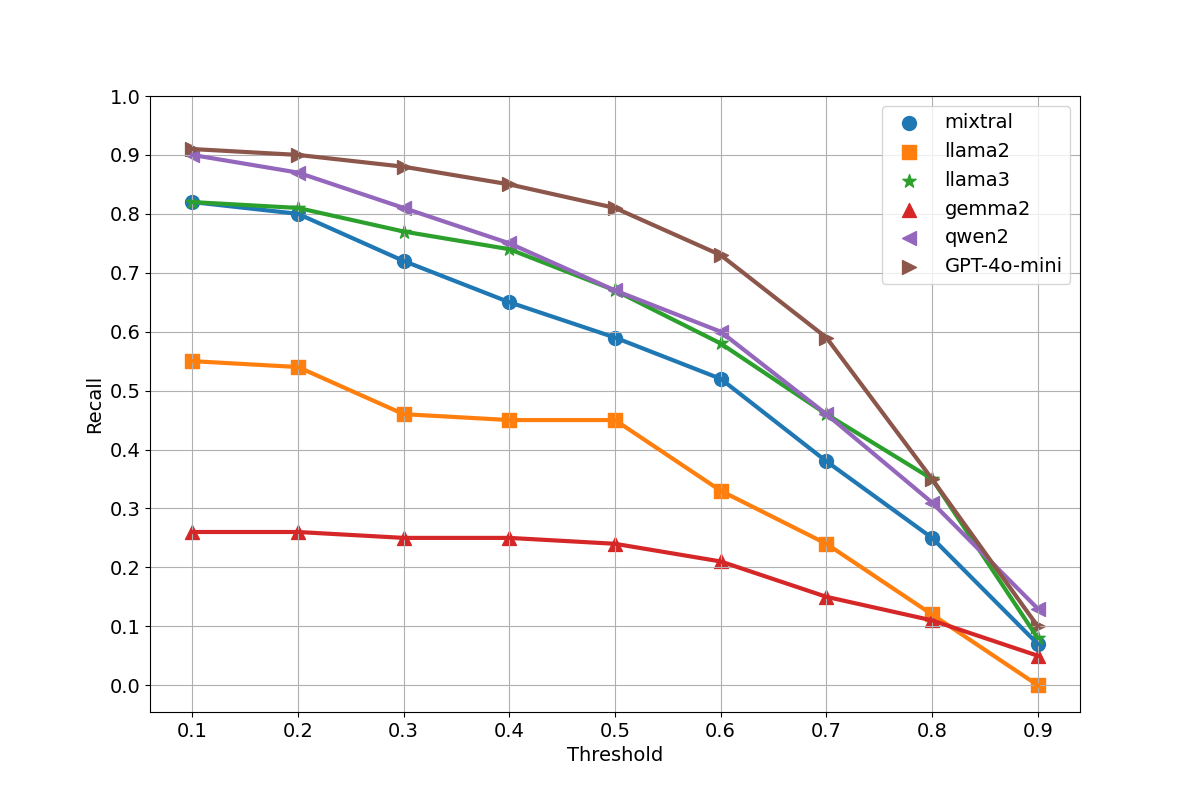}
         \caption{Recall}
         \label{fig:recall}
     \end{subfigure}
\caption{Comparison of Precision and Recall across different model thresholds.}
\end{figure}

For precision, Mixtral and LLaMA 3 show the highest precision scores across most threshold values, indicating they are better at minimizing false positives than other models, making them suitable for applications where minimizing false positives is crucial.
Qwen2 and GPT-4o-mini also perform well but with slightly lower precision than Mixtral and LLaMA 3.
Gemma2 maintains a relatively stable but lower precision score than the top-performing models, suggesting it produces more false positives.
  
For the recall, all models exhibit a decreasing trend in recall scores as the threshold increases, which is expected because higher thresholds typically result in fewer positive predictions.
GPT-4o-mini and Qwen2 are the best performers in terms of recall, making them suitable for applications where missing true positives is costly.
Gemma2 is consistently the weakest model in the recall, making it less suitable for tasks requiring high recall.

\section{Conclusion}
This study compares the performance of six large language models with GPT-4o on the SDG mapping task. The models evaluated are Mixtral, LLaMA 2, LLaMA 2, Gemma 2, Qwen2, and GPT-4o-mini. Since the dataset lacks ground truth labels, the output of GPT-4o (with a threshold of 0.5) is used as the baseline for comparison.
Generally, the F1 score, precision, and recall curves of the four models, GPT-4o-mini, Mixtral, LLaMA 3, and Qwen2, exhibit similar trends across thresholds ranging from 0.1 to 0.9. Their performances in the SDG mapping task are comparable. Conversely, the performances of LLaMA 2 and Gemma 2 are notably less satisfactory.
Among these, GPT-4o-mini performs slightly better, LLaMA 3 and Qwen2 are also reliable models.
Depending on the task requirements, selecting an appropriate threshold can optimize the balance between precision and recall. In this study, setting the threshold to 0.5 or 0.6 is advisable, as all the models' F1 scores and precision start to drop off beyond this point.
For tasks requiring high recall and balanced performance, GPT-4o-mini is recommended. Consider fine-tuning models like Mixtral, Qwen2, and LLaMA 3 for applications where precision is more critical.


\end{document}